\documentclass{article}

\usepackage{amsmath,amsfonts,bm}

\def\eqref#1{equation~\ref{#1}}
\def\1{\bm{1}}

\DeclareMathAlphabet{\mathsfit}{\encodingdefault}{\sfdefault}{m}{sl}
\SetMathAlphabet{\mathsfit}{bold}{\encodingdefault}{\sfdefault}{bx}{n}

\newcommand{\E}{\mathbb{E}}

\newcommand{\KL}{D_{\mathrm{KL}}}

\DeclareMathOperator*{\argmax}{arg\,max}

\usepackage{microtype}
\usepackage{graphicx}
\usepackage{subfigure}
\usepackage{booktabs} % for professional tables
\usepackage{nicefrac} % compact symbols for 1/2, etc.
\usepackage{url}
\usepackage{setspace}
\usepackage{amsthm}
\usepackage{amsmath}
\newtheorem{thm}{Theorem}
\newtheorem{assm}{Assumption}

\usepackage{mathtools}
\DeclarePairedDelimiter{\norm}{\lVert}{\rVert}
\usepackage{mathrsfs}
\usepackage{color}
\usepackage{stackengine}
\usepackage{esvect}
\usepackage{relsize}
\def\delequal{\mathrel{\ensurestackMath{\stackon[1pt]{=}{\scriptstyle\Delta}}}}
\newcommand{\code}[1]{\texttt{#1}}
\newcommand{\red}[1]{\textcolor{red}{\boldsymbol{#1}}}
\usepackage{algpseudocode}
\usepackage{algorithm}
\usepackage{fancyvrb}
\VerbatimFootnotes 
\usepackage[colorlinks]{hyperref}

\DeclareMathOperator*{\myargmax}{argmax}

\newtheoremstyle{named}{}{}{\itshape}{}{\bfseries}{}{.5em}{\thmnote{#3}}
\theoremstyle{named}
\newtheorem*{namedtheorem}{Theorem}

\usepackage[accepted]{icml2020}

\icmltitlerunning{Information-Theoretic Local Minima Characterization and Regularization}

\begin{document}

\twocolumn[
\icmltitle{Information-Theoretic Local Minima Characterization and Regularization}

\begin{icmlauthorlist}
\icmlauthor{Zhiwei Jia}{ucsd}
\icmlauthor{Hao Su}{ucsd}
\end{icmlauthorlist}

\icmlaffiliation{ucsd}{University of California, San Diego}
\icmlcorrespondingauthor{Zhiwei Jia}{zjia@ucsd.edu}
\icmlcorrespondingauthor{Hao Su}{haosu@eng.ucsd.edu}

\icmlkeywords{Deep Learning Theory, Regularization, Generalization, Local Minima}

\vskip 0.3in
]

\printAffiliationsAndNotice{}

\begin{abstract}
Recent advances in deep learning theory have evoked the study of generalizability across different local minima of deep neural networks (DNNs). While current work focused on either discovering properties of good local minima or developing regularization techniques to induce good local minima, no approach exists that can tackle both problems. We achieve these two goals successfully in a unified manner. Specifically, based on the observed Fisher information we propose a metric both strongly indicative of generalizability of local minima and effectively applied as a practical regularizer. We provide theoretical analysis including a generalization bound and empirically demonstrate the success of our approach in both capturing and improving the generalizability of DNNs. Experiments are performed on CIFAR-10, CIFAR-100 and ImageNet for various network architectures.
\end{abstract}

\section{Introduction}
Recently, there has been a surge in the interest of acquiring a theoretical understanding over deep neural network's behavior. Breakthroughs have been made in characterizing the optimization process, showing that learning algorithms such as stochastic gradient descent (SGD) tend to end up in one of the many local minima which have close-to-zero training loss \citep{choromanska2015loss, dauphin2014identifying, kawaguchi2016deep, nguyen2018optimization, du2018gradient}. However, these numerically similar local minima typically exhibit very different behaviors in terms of generalizability. It is, therefore, natural to ask two closely related questions: (a) What kind of local minima can generalize better? (b) How to find those better local minima? 

To our knowledge, existing work focused only on one of the two questions. For the ``what'' question, various definitions of ``flatness/sharpness'' have been introduced and analyzed \citep{keskar2016large, neyshabur2017pac, neyshabur2017exploring, wu2017towards, liang2017fisher}. However, they suffer from one or more of the problems: (1) being mostly theoretical with no or poor empirical evaluations on modern neural networks, (2) lack of theoretical analysis and understanding, (3) in practice not applicable to finding better local minima. Regarding the ``how'' question, existing approaches ~\citep{hochreiter1997flat, sokolic2017robust, chaudhari2016entropy, hoffer2017train, neyshabur2015path, izmailov2018averaging} share some of the common drawbacks: (1) derived only from intuitions but no specific metrics provided to characterize local minima, (2) no or weak analysis of such metrics, (3) not applicable or no consistent generalization improvement for modern DNNs. 

In this paper, we tackle both the ``what'' and the ``how'' questions in a unified manner. Our answer provides both the theory and applications for the generalization problems across different local minima. Based on the determinant of Fisher information estimated from the training set, we propose a metric that \emph{solves all the aforementioned issues}. The metric can well capture properties that characterize local minima of different generalization ability. We provide its theoretical analysis, primarily a generalization bound based on PAC-Bayes \citep{mcallester1999some, mcallester1999pac}. For modern DNNs in practice, it is necessary to provide a tractable approximation of our metric. We propose an intuitive and efficient approximation to compare it across different local minima. Our empirical evaluations fully illustrate the effectiveness of the metric as a strong indicator of local minima's generalizability. Moreover, from the metric we further derive and  design a practical regularization technique that guides the optimization process in finding better generalizable local minima. The experiments on image classification datasets demonstrate that our approach gives consistent generalization boost for a range of DNN architectures. Codes are available at \url{https://github.com/SeanJia/InfoMCR}.

\section{Related Work} \label{related_work}

It has been empirically shown that larger batch sizes lead to worse generalization \citep{keskar2016large}. \citet{hoffer2017train} analyzed how the training dynamics is affected by different batch sizes and presented a perturbed batch normalization technique for better generalization. While it effectively improves generalization for large-batch training, a specific metric that indicates the generalizability is missing. Similarly, \citet{elsayed2018large} employed a structured margin loss to improve performance of DNNs w.r.t. noise and adversarial attack yet no metric was proposed. Furthermore, this approach essentially provided no generalization gain in the normal training setup.

The local entropy of the loss landscape was proposed to measure ``flatness'' in \citet{chaudhari2016entropy}, which also designed an entropy-guided SGD that achieves faster convergence in training DNNs. However, the method does not consistently improve generalization, e.g., a decrease of performance on CIFAR-10 \citep{krizhevsky2009learning}. Another method that focused on modifying the optimization process is the Path-SGD proposed by \citet{neyshabur2015path}. Specifically, the authors derived an approximate steepest descent algorithm that utilizes the path-wise norm regularization to achieve better generalization. The authors only evaluated it on a two-layer neural network, very likely since the path norm is computationally expensive to optimize during training. 
 
A flat minimum search algorithm was proposed by \citet{hochreiter1997flat} based on the ``flatness'' of local minima defined as the volume of local boxes. Yet since the boxes have their axes aligned to the axes of the model parameters, their volumes could be significant underestimations of ``flatness'' for over-parametrized networks, due to the specific spectral density of Hessian of DNNs studied in \citet{pennington2018spectrum, sagun2017empirical}. The authors of \citet{wu2017towards} also characterized the ``flatness'' by volumes. They considered the inverse volume of the basin of attraction and proposed to use the Frobenius norm of Hessian at the local minimum as a metric. In our experiments, we show that their metric does not accurately capture the generalization ability of local minima under different scenarios. Moreover, they have not derived a  regularizer from their metric.

Based on a ``robustness'' metric, \citet{sokolic2017robust} derived a regularization technique that successfully improves generalization on multiple image classification datasets. Nevertheless, we show that their metric fails to capture the generalizability across different local minima.

By using the Bayes factor, \citet{mackay1992practical} studied the generalization ability of different local minima obtained by varying the coefficient of L2 regularization. It derived a formula involving the determinant of Hessian, similar to the one in ours. Whereas, this approach has restricted settings and, without proposing an efficient approximation, its metric is not applicable to modern DNNs, let alone serving as a regularizer. A generalization bound is missing in \citet{mackay1992practical} as well.

In a broader context of the ``what'' question, properties that capture the generalization of neural networks have been extensively studied. Various complexity measures for DNNs have been proposed based on norm, margin, Lipschitz constant, compression and robustness \citep{bartlett2002rademacher, neyshabur2015norm, sokolic2017robust, xu2012robustness, bartlett2017spectrally, zhou2018nonvacuous, dziugaite2017computing, arora2018stronger, jiang2018predicting}. While some of them aimed to provide tight generalization bounds and some of them to provide better empirical results, none of the above approaches explored the ``how'' question at the same time.

Very recently, \citet{karakida2019universal} and \citet{sun2019lightlike} studied the Fisher information of the neural network through the lens of its spectral density. In specific, \citet{karakida2019universal} applied mean-field theory to study the statistics of the spectrum and the appropriate size of the learning rate. Also, an information-theoretic approach, \citet{sun2019lightlike} derived a novel formulation of the minimum description length in the context of deep learning by utilizing tools from singular semi-Riemannian geometry. 

\section{Outline and Notations} \label{pform}
In a typical $K$-way classification setting, each sample $x \in \mathcal{X}$ belongs to a single class denoted $c_x \in \{0, 1, ..., K\}$ according to the probability vector $y \in \mathcal{Y}$, where $\mathcal{Y}$ is the k-dimensional probability simplex so that $p(c_x=i) = y_i$ and $\sum_i y_i = 1$. Denote a feed-forward DNN parametrized by $w \in \mathbb{R}^W$ as $f_w: ~\mathcal{X} \rightarrow \mathcal{Y}$, which uses nonlinear activation functions and a softmax layer at the end. Denote the cross entropy loss as $\ell(f_w(x), y) = - \sum_i y_i \ln f_w(x)_i$. Denote the training set as $\mathcal{S}$, defined over $\mathcal{X} \times \mathcal{Y}$ with $|\mathcal{S}| = N$. The training objective is given as $\mathcal{L}(\mathcal{S}, w) = \frac{1}{N}\sum_{(x, y) \sim \mathcal{S}} \ell(f_w(x), y)$. Assume $\mathcal{S}$ is sampled from some true data distribution denoted $\mathcal{D}$, we can define expected loss $\mathcal{L}(\mathcal{D}, w) = \mathop{\mathbb{E}}_{(x, y) \sim \mathcal{D}}[\ell(f_w(x), y)]$. Throughout this paper, we refer a local minimum of $\mathcal{L}(\mathcal{S}, w)$ corresponding to a local minimizer $w_0$ as just the local minimum $w_0$. Our paper's outline and main achievements are: 
\begin{itemize}
    \item In Sec. \ref{fisher} we relates Fisher information to neural network training as a prerequisite.
    \item In Sec. \ref{metric} we propose a metric $\gamma(w_0)$ that well captures local minima's generalizability.
    \item In Sec. \ref{gbound} we provide a generalization bound related to $\gamma(w_0)$.
    \item In Sec. \ref{appro} we propose an approximation $\widehat{\gamma}(w_0)$ for $\gamma(w_0)$, which is shown to be very effective in Sec. \ref{exp_char} via extensive empirical evaluations.
    \item In Sec. \ref{regularization} we devise a practical regularizer from $\widehat{\gamma}(w_0)$ that consistently improves generalizability across different DNNs, as evaluated in Sec. \ref{exp_reg}.
\end{itemize}

\subsection{Other Notations} \label{nota}
Denote $\nabla_w$ as gradient, $\mathbf{J}_w[\cdot]$ as Jacobian matrix, $\nabla_w^2$ as Hessian, $\KL(\cdot \Vert \cdot)$ as KL divergence, $\norm{\cdot}_2$ as spectrum or Euclidean norm, $\norm{\cdot}_F$ as Frobenius norm, $|\cdot|$ as determinant, $\textrm{tr}(\cdot)$ as trace norm, $\rho(\cdot)$ as spectral radius, $\ell\ell_{\mathcal{S}}(w)$ as log-likelihood on $\mathcal{S}$, and $[\cdot]_i$ for selecting the $i^{\rm{th}}$ entry.

We define $\boldsymbol{\ell}_x(w) \in \mathbb{R}^K$ whose $i^{\rm{th}}$ entry is $-\ln f_w(x)_i$ so that $\ell(f_w(x), y) = \boldsymbol{\ell}_x(w)^T y$. We define $\tilde{y} \in \mathbb{R}^K$ as the one-hot version of $y$, i.e., only keep the largest dimension as 1. Then we define $\tilde{\mathcal{L}}(\mathcal{S}, w) \in \mathbb{R}^N$ as the one-hot and vectorized version of $\mathcal{L}(\mathcal{S}, w)$, i.e., a vector whose entries are $\ell(f_w(x), {\tilde{y}})$ for $(x, y) \in \mathcal{S}$.
In other words, we approximate the cross entropy loss $\ell(f_w(x), y)$ by $\ell(f_w(x), \tilde{y})$. 

\section{Local Minimum and Fisher Information} \label{fisher}
First of all, if $y$ is strictly one-hot and the training accuracy achieved at $w_0$ is 100\%, then $w_0$ cannot be a local minimizer, because the cross entropy loss remains positive even if arbitrarily close to zero. To admit local minima of full training accuracy, we assume the widely used label smoothing (LS) \citep{szegedy2016rethinking} is applied to train all models in our analysis. LS enables us to assume a local minimum $w_0$ of the training loss with $\sum_{(x, y) \in \mathcal{S}} \KL(f_{w_0}(x) \Vert y) = 0$. Although empirically we find that both our proposed metric and derived regularizer work similarly well without LS.

With LS in mind, each sample $(x, y) \in \mathcal{S}$ has its label $c_x$ sampled by $p(c_x=~i|x) = y_i$, denoted as $c_x  \sim y$. We denote the training data distribution as $(x, c_x) \sim \mathcal{S}$. The joint probability $p(x, c_x)$ modeled by the DNN is $p(x, c_x=i; w) = p(c_x=~i | x; w)\ p(x) = [f_w(x)]_i\ p(x)$ with $p(x) = \frac{1}{N}$. We can relate the training loss $\mathcal{L}(\mathcal{S}, w)$ to the negative log-likelihood $- \ell\ell_{\mathcal{S}}(w)$ by:
\begin{align*} \label{eq:1}
\mathcal{L}(\mathcal{S}, w) &= \frac{1}{N} \sum_{(x, y) \in \mathcal{S}} \boldsymbol{\ell}_x(w)^T y \\
&= - \frac{1}{N}\sum_{(x, y) \in \mathcal{S}} \mathop{\mathbb{E}}_{c_x \sim y} \ln p(c_x|x; w) \\
&= - \frac{1}{N} \ell\ell_{\mathcal{S}}(w) + \ln\frac{1}{N}
\end{align*}
\vspace{-20px}
\[\textrm{where}\ - \ell\ell_{\mathcal{S}}(w) = - \sum_{(x, y) \in \mathcal{S}} \mathbb{E}_{c_x \sim y} \ln p(x, c_x; w)\]
Also, $w_0$ corresponds to a local maximum of the likelihood function. The observed Fisher information \citep{efron1978assessing} evaluated at $w_0$ is defined using the Hessian of the negative log-likelihood, i.e.,
\begin{align} \label{eq:b}
    \mathcal{I}_{\mathcal{S}}(w_0) &= - \frac{1}{N} \nabla_w^2 \ell\ell_{\mathcal{S}}(w_0) =  \nabla_w^2 \mathcal{L}(\mathcal{S}, w_0) \nonumber \\
    &= \mathlarger{\mathop{\mathbb{E}}}_{(x, c_x) \sim \mathcal{S}} [\nabla_w \ln p_{w_0}(c_x) \nabla_w \ln p_{w_0}(c_x)^T]
\end{align}
where $p_{w_0}(c_x)$ denotes $p(c_x|x;w_0)$. The first equality is straightforward; the second has its proof in Appendix A. Since $p(c_x=i|x) = y_i$ and $\ln p(c_x =~ i|x; w_0) = [\boldsymbol{\ell}_x(w_0)]_i$, we can further simplify the Equation \ref{eq:b} to:
\begin{equation} \label{eq:1}
\mathcal{I}_{\mathcal{S}}(w_0) = \frac{1}{N}\sum_{(x, y) \in \mathcal{S}} \sum_{i=1}^K \nabla_{w}[\boldsymbol{\ell}_x(w_0)]_i \nabla_{w}[\boldsymbol{\ell}_x(w_0)]_i^T
\end{equation}
\textbf{Remark}: A global minimum $w_0$, if exists, is equivalent to a local minimum with 100\% training accuracy. At such $w_0$, we have $\nabla_w \ell(f_{w_0}(x), y) = \boldsymbol{0}$ as $\KL(f_{w_0}(x) \Vert y) = 0$; however, we also have $\mathcal{I}_{\mathcal{S}}(w_0) \in \mathbb{R}^{W\times W} \neq \boldsymbol{0}$. 

\section{Local Minima Characterization} \label{charactering}

In this section, we derive and propose our metric, provide a PAC-Bayes generalization bound, and lastly, propose and give intuitions of an effective approximation of our metric for modern DNNs. 

\subsection{Fisher Determinant as Generalization Metric} \label{metric}
We would like a metric to compare different local minima.
Under the Assumption \ref{assume_unique}, we can partition the parameter space of the neural network $f_w$ into disjoint regions, each is a small neighborhood of a local minimum taken into account. Formally, for a local minimum $w_0$ and a sufficiently small $V > 0$, we define the model class $\mathcal{M}(w_0)$ as the largest connected subset of $\{w \in ~\mathbb{R}^W: \mathcal{L}(\mathcal{S}, w) \leq h \}$ that contains $w_0$, where the height $h$ is defined as a real number such that the volume (namely the Lebesgue measure) of $\mathcal{M}(w_0)$ is $V$. By the Intermediate Value Theorem, for any sufficiently small $V$ there exists a corresponding height $h$. In essence, a local minimum $w_0$ of the entire parameter space becomes the global minimum of the model class $\mathcal{M}(w_0)$.

Formulated as a model class selection problem, we can compare different local minima by comparing their associated model classes. 
We propose our metric $\gamma(\cdot)$, where lower $\gamma(w_0)$ indicates a better generalizable local minimum $w_0$:
\begin{equation}
\gamma(w_0) = \ln |\mathcal{I}_{\mathcal{S}}(w_0)|
\end{equation}
As a metric, $\gamma(w_0)$ requires $|\mathcal{I}_{\mathcal{S}}(w_0)|\ \neq 0$. Therefore, we state the following Assumption \ref{assume_unique}.

\begin{assm} \label{assume_unique}
The local minima $w_0$ we care about in the comparison are well isolated and unique in their corresponding neighborhood $\mathcal{M}(w_0)$.
\end{assm}

The Assumption \ref{assume_unique} is quite reasonable. For state-of-the-art network architectures used in practice, this is often the fact. To be precise, the Assumption \ref{assume_unique} is violated when the Hessian matrix at a local minimum is singular. Specifically, \citet{orhan2018skip} summarizes three sources of the singularity: (i) due to a dead neuron, (ii) due to identical neurons, and (iii) linear dependence of the neurons. As well demonstrated in \citet{orhan2018skip}, network with skip connection, e.g. ResNet \citep{he2016deep}, WRN \citep{wideresnet}, and DenseNet \citep{huang2017densely} used in our experiments, can effectively eliminate all the aforementioned singularity. % somehow we need an extra blank line here% 

In \citet{dinh2017sharp}, the authors pointed out another source of the singularity specifically for networks with scale-invariant activation functions, e.g. ReLU. Namely, one can rescale the model parameters layer-wise so that the underlying function represented by the network remains unchanged in the region. In practice, this issue is not critical. Firstly, most modern deep ReLU networks, e.g. ResNet, WRN, and DenseNet, have normalization layers, e.g. BatchNorm \citep{ioffe2015batch}, applied before the activations. BatchNorm shifts all the inputs to the ReLU function, equivalently shifting the ReLU horizontally which makes it no longer scale-invariant. Secondly, due to the ubiquitous use of Gaussian weights initialization scheme and weight decay, most local minima obtained by gradient learning have weights of a relatively small norm. Consequently, in practice, we will not compare two local minima essentially the same but have one as the rescaled version of the other with a much larger norm of the weights. 

Note that normally we have a limited size of the dataset, and so an approximation of $\gamma(w_0)$ is a must. We present our approximation scheme and its intuition in Sec. \ref{appro}.

\subsubsection{Connection to Fisher Information Approximation (FIA) Criterion}

Our metric $\gamma(w_0)$ is closely related to the FIA criterion. Based on the MDL principle \citep{rissanen1978modeling}, \citet{rissanen1996fisher} derived the FIA criterion to compare statistical models. Tailored to our setting, each model class $\mathcal{M}(w_0)$ has its FIA criterion as (lower FIA is better):
\begin{align*}
    \textrm{FIA} = &-\sum_{(x, y) \in \mathcal{S}} \mathlarger{\mathop{\mathbb{E}}}_{c_x \sim y} \ln p(x, c_x; w_0) \\
     &+ \frac{W}{2}\ln\frac{N}{2\pi} + \ln \int_{\mathcal{M}(w_0)} \sqrt{|\mathcal{J}(w)|}\ dw
\end{align*}
Where $\mathcal{J}(w)$ is the expected Fisher information evaluated at $w$. Notice that all regularity conditions of the FIA criterion are satisfied for the local minimum $w_0$ (also the global optimum of the model class), provided 100\% training accuracy and the Assumption \ref{assume_unique}. Ignoring the constant terms and assuming the training loss is locally quadratic in $\mathcal{M}(w_0)$ (later formalized and validated as Assumption \ref{assume}), the RHS becomes $\ln V + \frac{1}{2}\ln |\mathcal{J}(w_0)|$. Remind that $V$ is defined as the volume of $\mathcal{M}(w_0)$, also a constant.

Essentially in our metric we use the observed Fisher information in place of the expected one, making our metric tractable and applicable to modern DNNs.

\subsubsection{ Connection to Existing Flatness/Sharpness Metrics}

As mentioned in Sec. \ref{related_work}, the ``flatness'' of a local minimum was firstly related to the generalization ability of the neural network in \citet{hochreiter1997flat}, where the concept and the method are both preliminary. The idea is recently popularized in the context of deep learning by a series of paper such as \citet{keskar2016large, chaudhari2016entropy, wu2017towards}. Our approach roughly shares the same intuition with these existing works, namely, a ``flat'' local minimum admits less complexity and so generalizes better than a ``sharp'' one. To our best knowledge, our paper is the first among these work that provides both the theoretical analysis including a generalization bound and the empirical verification of both an efficient metric and a practical regularizer for modern network architectures. 

\subsection{Generalization Bound} \label{gbound}

\begin{assm} \label{assume}
Given the training loss $\mathcal{L}(\mathcal{S}, w)$, its local minimum $w_0$ satisfying Assumption \ref{assume_unique} and the associated neighborhood $\mathcal{M}(w_0)$ whose volume $V$ is sufficiently small, as described in Sec. \ref{pform}, \ref{fisher} and \ref{metric}, respectively, when confined to $\mathcal{M}(w_0)$, we assume that $\mathcal{L}(\mathcal{S}, w)$ is quadratic. 
\end{assm}

The Assumption \ref{assume} is quite reasonable as well. \citet{grunwald2007minimum} suggests that, a log-likelihood function, under regularity conditions (1) existence of its $1^{\rm{st}}$, $2^{\rm{nd}}$ \& $3^{\rm{rd}}$ derivatives and (2) uniqueness of its maximum in the region, behaves locally like a quadratic function around its maximum. In our case, $\mathcal{L}(\mathcal{S}, w)$ corresponds to the log-likelihood function $\ell\ell_{\mathcal{S}}(w)$ and so $w_0$ corresponds to a local maximum of $\ell\ell_{\mathcal{S}}(w)$. Since $\mathcal{L}(\mathcal{S}, w)$ is analytic and $w_0$ is the only local minimum of $\mathcal{L}(\mathcal{S}, w)$ in $\mathcal{M}(w_0)$, the training loss indeed can be considered locally quadratic.

Similar to \citet{langford2002not}, \citet{harvey2017nearly} and \citet{neyshabur2017exploring}, we apply the PAC-Bayes Theorem \citep{mcallester2003simplified} to derive a generalization bound for our metric. Specifically, we pick a uniform prior $\mathcal{P}$ over $w \in \mathcal{M}(w_0)$ according to the maximum entropy principle and pick the posterior $\mathcal{Q}$ of density $q(w) \propto e^{-|\mathcal{L}_0 - \mathcal{L}(\mathcal{S}, w)|}$ with $\mathcal{L}_0 \delequal \mathcal{L}(\mathcal{S}, w_0)$. Then Theorem \ref{main} bounds the expected generalization loss using $\gamma(w_0)$ (proved in Appendix B).

\begin{thm} \label{main}
Given $|\mathcal{S}| = N$, $\mathcal{D}$, $\mathcal{L}(\mathcal{S}, w)$ and $\mathcal{L}(\mathcal{D}, w)$ described in Sec. \ref{pform}, a local minimum $w_0$, the volume $V$ of $\mathcal{M}(w_0)$ sufficiently small, the Assumption \ref{assume_unique} \& \ref{assume} satisfied, and $\mathcal{P}, \mathcal{Q}$ defined above, for any $\delta \in (0, 1]$, we have with probability at least $1 - \delta$ that: 
\begin{align*}
\mathop{\mathbb{E}}_{w \sim \mathcal{Q}} [\mathcal{L}(\mathcal{D}, w)] &\leq \mathop{\mathbb{E}}_{w \sim \mathcal{Q}} [\mathcal{L}(\mathcal{S}, w)] + 2 \sqrt{ \frac{2 \mathcal{L}_0 + 2 \mathcal{A} + \ln \frac{2N}{\delta}}{N - 1}} \\
\textrm{where}\ \mathcal{A} &= \frac{1}{4\pi e}W V^{\frac{2}{W}} \pi^{\frac{1}{W}} \exp\{\frac{\red{\gamma(w_0)}}{W}\}
\end{align*}

\end{thm}

Where $W$ is the number of model parameters (defined in Sec. \ref{pform}) and
$V$ the volume controlling the size of the neighborhood taken into account around $w_0$ (defined in Sec. \ref{metric}).
In short, Theorem \ref{main} shows that a lower $\gamma(w_0)$ indicates a local minimum $w_0$ of better generalization.

\subsection{Approximation} \label{appro}
As stated in Sec. \ref{fisher}, in practice an approximation of $\gamma(w_0)$ as $\widehat{\gamma}(w_0)$ is necessary, as calculating $\gamma(w_0)$ involves computing the product of all $W$ non-zero eigenvalues of the Fisher information matrix. Assume an imagined training set $\mathcal{S}'$ of size $W$ and a local minimum $w_0$ of $\mathcal{L}(\mathcal{S}', w)$; then $\ln |\mathcal{I}_{\mathcal{S'}}(w_0)|$ is well defined on the full-rank Fisher information denoted as $\mathcal{I}_{\mathcal{S'}}(w_0)$. In reality, we only have a training set $\mathcal{S} \subset \mathcal{S}'$ with $|\mathcal{S}|$ non-zero eigenvalues of the singular matrix $\mathcal{I}_{\mathcal{S}}(w_0)$. 
% Notice that $w_0$ is also a local minimum of $\mathcal{L}(\mathcal{S}, w)$ since $\sum_{(x, y) \in \mathcal{S'}} \KL(f_{w_0}(x) \Vert y) = 0$ as assumed in Sec. \ref{fisher}.
Similar to the approach in \citet{karakida2019universal}, we propose to approximate eigenvalues of $\mathcal{I}_{\mathcal{S'}}(w_0)$ by the non-zero eignevalues of $\mathcal{I}_{\mathcal{S}}(w_0)$, or equivalently, as shown later, by the eigenvalues of sub-matrices of $\mathcal{I}_{\mathcal{S'}}(w_0)$.

First of all, we replace $y$ by its one-hot version $\tilde{y}$ defined in Sec. \ref{nota}, drastically reducing the cost of gradient calculation. This is reasonable since $y$ and $\tilde{y}$ are very close. With $\tilde{\mathcal{L}}(\mathcal{S}, w) \in \mathbb{R}^N$ defined in Sec. \ref{nota}, according to Equation \ref{eq:1}, we have $\mathcal{I}_{\mathcal{S'}}(w_0) \in \mathbb{R}^{W \times W}$ as:
\begin{align} \label{eq:one_hot_approx}
\mathcal{I}_{\mathcal{S'}}(w_0) &\approx \frac{1}{W}\sum_{(x, y) \in \mathcal{S}'} \nabla_{w}[\boldsymbol{\ell}_x(w_0)]_{\textbf{y}} \nabla_{w}[\boldsymbol{\ell}_x(w_0)]_{\textbf{y}}^T \nonumber \\
&\quad\ \textrm{where}\ \textbf{y} = \myargmax(y) \nonumber \\
&= \frac{1}{W} \mathbf{J}_{w}[\tilde{\mathcal{L}}(\mathcal{S'}, w)]^T \mathbf{J}_{w}[\tilde{\mathcal{L}}(\mathcal{S'}, w)] \nonumber \\
&= \frac{1}{W} \mathbf{J}_{w}[\tilde{\mathcal{L}}(\mathcal{S'}, w)]\ \mathbf{J}_{w}[\tilde{\mathcal{L}}(\mathcal{S'}, w)]^T 
\end{align}
Let $\{\lambda_m\}_{m=1}^W$ denote the eigenvalues of $\mathcal{I}_{\mathcal{S'}}(w_0)$; then $\gamma(w_0) = \ln \prod_{m = 1}^W \lambda_m = \sum_{m = 1}^W \ln \lambda_m$. Ideally we want to perform a Monte-Carlo estimation of $\gamma(w_0)$ by randomly sampling $N' < N \ll W$ eigenvalues from $\{\lambda_m\}_{m=1}^W$, where $N$ is the size of $\mathcal{S}$. We denote the samples as $\{\lambda_n\}_{n=1}^{N'}$ and we have $\frac{W}{N'}\sum_{n=1}^{N'} \ln \lambda_n \approx \sum_{m=1}^W \ln \lambda_m$. Suppose the estimation is run $T$ times, we have $\lim_{T \rightarrow \infty} \frac{1}{T} \sum_{t=1}^T \frac{W}{N'}\sum_{n=1}^{N'} \ln \lambda_n = \gamma(w_0)$.

Then the eigenvalue approximation comes in. We sample $\mathcal{S}^t \subset \mathcal{S}$ i.i.d. with $|\mathcal{S}^t| = N'$ for $T$ times and define 
\vspace{1pt}
\begin{equation}
\xi^t(w_0) \delequal \mathbf{J}_{w}[\tilde{\mathcal{L}}(\mathcal{S}^t, w_0)] \mathbf{J}_{w}[\tilde{\mathcal{L}}(\mathcal{S}^t, w_0)]^T \in \mathbb{R}^{N' \times N'}
\end{equation}
Notice that $\xi^t(w_0)$ is a principal sub-matrix of $W \mathcal{I}_{\mathcal{S'}}(w_0)$ by removing rows \& columns for data in $\mathcal{S}$ \textbackslash\ $\mathcal{S}^t$. According to Theorem 2 in Appendix C and properties of the spectral density of Fisher information \citep{pennington2018spectrum, sagun2017empirical, karakida2019universal}, one can well approximate the eigenvalues of $\mathcal{I}_{\mathcal{S'}}(w_0)$ by those of its sub-matrices. Therefore we
define the estimation $\widehat{\gamma}(w_0)$ as:
% propose to estimate $\gamma(w_0)$ by defining $\widehat{\gamma}(w_0)$ and derive: 
\begin{equation}
    \widehat{\gamma}(w_0) \delequal \frac{1}{T} \sum_{t = 1}^{T} \ln \big\vert\xi^t(w_0)\big\vert
\end{equation}
The relation between $\widehat{\gamma}(w_0)$ and $\gamma(w_0)$ is given as:
\begin{equation} \label{eq:2}
\quad \gamma(w_0) \approx \frac{W}{N'} \widehat{\gamma}(w_0) + W \ln \frac{1}{W}\ \ \textrm{as}\ \ T \rightarrow \infty \nonumber 
\end{equation} 

We leave the derivation of Equation \ref{eq:2} to Appendix C. In proposing $\widehat{\gamma}(w_0)$, we ignore the constants and irrelevant scaling factors. Empirically we find that given relatively large number of sample trials $T$, our metric $\widehat{\gamma}(\cdot)$ can effectively capture the generalizability of a local minimum even for a small $N'$ (details in Sec. \ref{exp_char} and Appendix D).

\section{Local Minima Regularization} \label{regularization}

Besides pragmatism, devising a practical regularizer based on $\gamma(w_0)$ also ``verifies'' our theoretical understanding of DNN training, helping the future improvement of the learning algorithms. Following the approximation scheme in Sec. \ref{appro}, it is natural to regularize $\gamma(w_0)$ during mini-batch learning by minimizing the product of $|\mathcal{B}|$ non-zero eigenvalues of the Fisher information computed $\mathcal{I}_{\mathcal{B}}(w_0)$, computed via the current batch $\mathcal{B}$, other than directly minimizing $\gamma(w_0)$.
However, this is far from practical due to the computation burden of:
\begin{enumerate} 
    \item computing the eigenvalues in each training step
    \item computing second-order derivatives (i.e., computing the gradients of $\widehat{\gamma}(w_0)$ with respect to $w_0$)
\end{enumerate}
There is another major challenge. All of our theoretical analysis of $\gamma(\cdot)$ works on the grounds that the Assumption \ref{assume_unique} \& \ref{assume} are reasonable and satisfied, i.e., the largest $|\mathcal{B}|$ eigenvalues of $\mathcal{I}_{\mathcal{B}}(w_0)$ evaluated at the local minimum $w_0$ are non-zero. However, directly minimizing the product of these positive eigenvalues pays too much attention to the smallest eigenvalues, which can easily result in zero eigenvalues, raising singularity and thus violating the assumptions. Instead, we need the effort more spread out. A good choice is to minimize the trace norm $\textrm{tr}\big(\mathcal{I}_{\mathcal{B}}(w_0)\big)$, which provides an upper bound of the product of eigenvalues in the form of: 
\[\prod_i \lambda_i \big(\mathcal{I}_{\mathcal{B}}(w_0)\big)^{1/|\mathcal{B}|} \leq \frac{1}{|\mathcal{B}|}\textrm{tr}\big(\mathcal{I}_{\mathcal{B}}(w_0)\big)\]
Although this bound is not be tight in general, we are tightening it when we minimize the trace norm. According to Corollary 1 in \citet{rodin2017variance}, we have:
\[
\frac{1}{|\mathcal{B}|}\textrm{tr}\big(\mathcal{I}_{\mathcal{B}}(w_0)\big) - \prod_i \lambda_i \big(\mathcal{I}_{\mathcal{B}}(w_0)\big)^{1/|\mathcal{B}|} \leq \sqrt{|\mathcal{B}| - 1}\ \sigma
\]
Where $\sigma$ denotes the standard deviation of the eigenvalues of $\mathcal{I}_{\mathcal{B}}(w_0)$. As pointed out in \citet{pennington2018spectrum, sagun2017empirical, karakida2019universal}, these eigenvalues are highly concentrated with only a few very large ``outliers'' which contribute the most to the variance. When we minimize the trace norm, i.e. the L1 norm of the eigenvalues, the largest few eigenvalues bear the most weight before they are reduced to a level that has the bound effectively tightened. Furthermore, computing the trace norm does not require computing eigenvalues; thus optimizing them removes the first computation burden.

Similar to the approach in Equation \ref{eq:one_hot_approx}, we approximate $y$ by its one-hot version $\tilde{y}$.
For simplicity, for the rest of this section, we denote $\tilde{y}$ as $y$ and correspondingly denote $\E_j [\tilde{\mathcal{L}}(\cdot, w_0)]_j$ as $\mathcal{L}(\cdot, w_0)$, where $\tilde{\mathcal{L}}$ is the one-hot vectorized loss defined in Sec. \ref{nota}.

Given a vector $x$, we have $\textrm{tr}(xx^T) = ||x||^2_2$. Therefore, we choose to approximate the trace norm as:
\begin{align*}
\textrm{tr}\big(\mathcal{I}_{\mathcal{B}}(w_0)\big) &\approx \frac{1}{|\mathcal{B}|} \sum_{(x, y) \in \mathcal{B}} \big\Vert\nabla_w \ell(f_{w_0}(x), y)\big\Vert^2_2
\end{align*}
To compute such quantity we need gradients for each individual data point.
We simplify this computation by grouping data points into batches and computing averaged gradients instead.
We randomly split $\mathcal{B}$ into $M$ sub-batches of equal size, namely $\{\mathcal{B}_i\}_{i=1}^{M}$. We define
\[\boldsymbol{g}_i \delequal \nabla_w \mathcal{L}(\mathcal{B}_i, w_0)\]
and then choose to optimize $\frac{1}{M}\sum_i^M \Vert\boldsymbol{g}_i\Vert^2_2$ instead of optimizing $\frac{1}{|\mathcal{B}|}\sum_{(x,y)\in\mathcal{B}}\Vert \nabla_w \ell(f_{w_0}(x), y) \Vert^2_2$, which drastically boosts the speed performance.

\begin{figure}[ht]
  \vspace{-10px}
  \centering
  \includegraphics[width=\columnwidth]{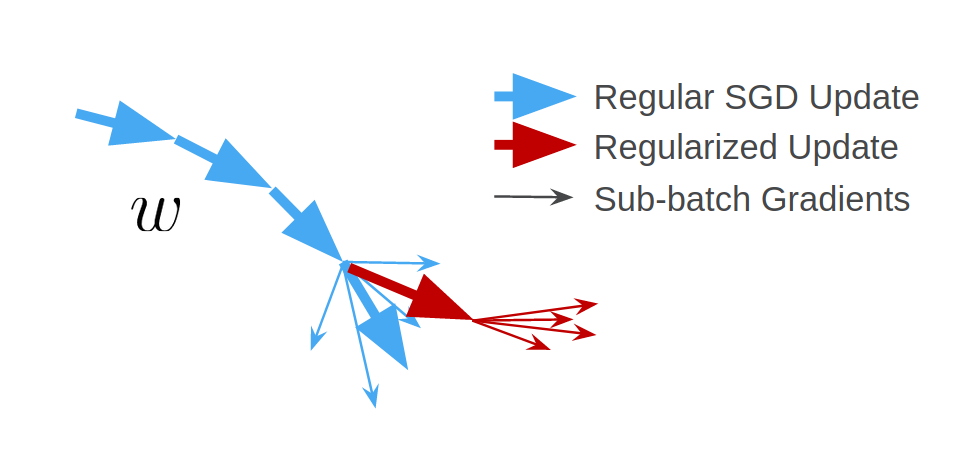}
   \vspace{-20px}
  \caption{An illustration of Algorithm \ref{reg_algo}. In essence, the regularizer guides the optimization process to areas with less divergent gradients of different data points within a mini-batch.}
  \label{fig:reg_illustrate}
\end{figure}

We deal with the second computation burden by adopting first order approximation. For any $w$, with a sufficiently small $\alpha > 0$, we have:
\[\tilde{\mathcal{L}}(\mathcal{B}_i, w - \alpha \boldsymbol{g}_i) \approx \tilde{\mathcal{L}}(\mathcal{B}_i, w) - \mathbf{J}_{w}[\tilde{\mathcal{L}}(\mathcal{B}_i, w)]\ \alpha \boldsymbol{g}_i\]
Thereby, we can estimate $\big\Vert\boldsymbol{g}_i\big\Vert^2_2$ by:
\begin{align*}
\alpha \big\Vert\boldsymbol{g}_i\big\Vert^2_2 &=  \frac{1}{|\mathcal{B}_i|}\sum_{j=1}^{|\mathcal{B}_i|}\Big[\mathbf{J}_{w}[\tilde{\mathcal{L}}(\mathcal{B}_i, w)]\ \alpha \boldsymbol{g}_i\Big]_j  \\
&\approx \frac{1}{|\mathcal{B}_i|}\sum_{j=1}^{|\mathcal{B}_i|}[\tilde{\mathcal{L}}(\mathcal{B}_i, w) - \tilde{\mathcal{L}}(\mathcal{B}_i, w - \alpha \boldsymbol{g}_i)]_j \\
&= \mathcal{L}(\mathcal{B}_i, w) - \mathcal{L}(\mathcal{B}_i, w - \alpha \boldsymbol{g}_i) \\
\end{align*}
Therefore, we propose to optimize the following regularized training objective for each mini-batch gradient descent step:
\begin{align} \label{eq:reg}
&\quad \mathcal{L}(\mathcal{B}, w) + \beta \mathcal{R}_{\alpha}(w)\ \ \textrm{where} \\ \mathcal{R}_{\alpha}(w) &\delequal \frac{1}{M} \sum_{i=1}^M \big[\mathcal{L}(\mathcal{B}_i, w) - \mathcal{L}(\mathcal{B}_i, w - \alpha \boldsymbol{g}_i)\big] \nonumber \\
&= \mathcal{L}(\mathcal{B}, w) - \frac{1}{M} \sum_{i=1}^M \mathcal{L}(\mathcal{B}_i, w - \alpha \boldsymbol{g}_i) \nonumber
\end{align}
Illustrated in Fig. \ref{fig:reg_illustrate}, an intuition is that Eq. \ref{eq:reg} penalizes a divergent set of gradients across samples in a mini-batch.

We omit any second order term when computing $\nabla_w \mathcal{R}_{\alpha}(w)$, simply by not back-propagating the gradient through $\boldsymbol{g}_i$. 
% As in gradient-based training we can never exactly reach the local minimum $w_0$, we optimize $\textrm{tr}\big(\mathcal{I}_{\mathcal{B}}(w)\big)$ during the entire training process.
% We also find that gradient clipping for the regularized loss, especially at the beginning of the training, is necessary to make the generalization boost consistent.
We outline our regularized training step as Algorithm \ref{reg_algo}, which has 3 hyper-parameters: $\alpha$, $\beta$ and $M$.

\begin{algorithm}
\caption{Regularized Gradient Descent \footnotemark[1]} \label{reg_algo}
\begin{algorithmic}[1]
\Procedure{Update}{$w, \mathcal{B}$; $\alpha, \beta, M$} 
% \Comment{Last 4 are hyper-parameters}
\State $\{\mathcal{B}_i\}_{i=1}^M \gets \mathcal{B}$ \Comment{Split the mini-batch $\mathcal{B}$}
\For{$i \gets 1$ to $M$}
    \State $\boldsymbol{g}_i \gets  \nabla_w \mathcal{L}(\mathcal{B}_i, w_0)$ 
    \State $\boldsymbol{g}_i \gets \textrm{copy}(\boldsymbol{g}_i)$ \Comment{Stop the gradient\footnotemark[2]}
\EndFor
\State $\mathcal{R}_{\alpha}(w) \gets \frac{1}{M} \sum_{i=1}^M \big[\mathcal{L}(\mathcal{B}_i, w) - \mathcal{L}(\mathcal{B}_i, w - \alpha \boldsymbol{g}_i)\big]$
\State $\nabla_w \mathcal{L}_{\textrm{reg}} \gets \nabla_w [\mathcal{L}(\mathcal{B}, w) + \beta \mathcal{R}_{\alpha}(w)]$
% \State Clip\footnotemark[3] the gradient $\nabla_w \mathcal{L}_{\textrm{reg}}$ with threshold $\tau$
% \Comment{}}
\State Update weights $w$ with $\nabla_w \mathcal{L}_{\textrm{reg}}$ 
% \Comment{Update with any gradient-based optimizer}
\EndProcedure
\end{algorithmic}
\end{algorithm}

\footnotetext[1]{Compatible with any gradient descent-based optimizer.}
\footnotetext[2]{Implemented as \verb|stop_gradient| in TensorFlow.}
% \footnotetext[3]{Implemented as \verb|clip_by_global_norm| in TensorFlow.}

\section{Experiments} \label{exp}
We perform two sets of experiments to illustrate the effectiveness of our metric $\gamma(w_0)$. We demonstrate that: (1) the approximation $\widehat{\gamma}(w_0)$ captures the generalizability well across local minima; (2) our regularization technique based on $\gamma(w_0)$ provides consistent generalization gain for DNNs.

Throughout our theoretical analysis, we assume that label smoothing (LS) is applied during model training in order to obtain well-defined local minima (first mentioned in Sec. \ref{fisher}). In all our empirical evaluations, we perform both the version with LS applied and without. Results are very similar and so we stick to the version without LS to be consistent with the original setup in papers of the various DNNs that we used. 
As a result, $\tilde{y}$ and $y$ refers to the same quantity.

\subsection{Experiments on Local Minima Characterization} \label{exp_char}

We perform comprehensive evaluations to compare our metric $\widehat{\gamma}(\cdot)$ with several others on ResNet-20 \citep{he2016deep} for the CIFAR-10 dataset (architecture details in Appendix E). Our metric consistently outperforms others in indicating local minima's generalizability. Specifically, \citet{sokolic2017robust} proposed a robustness-based metric used as a regularizer; \citet{wu2017towards} proposed to use Frobenius norm of the Hessian as a metric; \citet{keskar2016large} proposed a metric closely related to the spectral radius of Hessian. In summary, we compare 4 metrics, all evaluated at a local minimum $w$ given training set $\mathcal{S}$. All four metrics go for ``smaller values indicate better generalization''.
\begin{itemize}
  \item Robustness: $\frac{1}{N} \sum_{(x,y) \in \mathcal{S}} \big\Vert\textbf{J}_{x}[f_w(x)]\big\Vert^2_2$
  \item Frobenius norm: $\big\Vert \nabla^2_w \mathcal{L}(\mathcal{S}, w)\big\Vert_F^2$ 
  \item Spectral radius: $\rho( \nabla^2_w \mathcal{L}(\mathcal{S}, w))$
  \item Ours: $\widehat{\gamma}(w) =\frac{1}{T} \sum_{t = 1}^{T} \ln |\xi(\mathcal{S}^t, w_0)|$, $\ \mathcal{S}^t \subset \mathcal{S}$
\end{itemize}
Both the Frobenius norm and the spectral radius based metric are related to ours, as from Equation \ref{eq:b} we have $\big\Vert \nabla^2_w \mathcal{L}(\mathcal{S}, w)\big\Vert_F^2 = \big\Vert\mathcal{I}_{\mathcal{S}}(w)\big\Vert_F^2$ and $\rho(\nabla^2_w \mathcal{L}(\mathcal{S}, w)) = \rho(\mathcal{I}_{\mathcal{S}}(w))$. These two metric, however, are too expensive to compute for the entire training set $\mathcal{S}$; we instead calculate them by averaging the results for $T$ sampled $\mathcal{S}^t \subset \mathcal{S}$, similar to when we compute $\widehat{\gamma}(w)$. We leave details of how we exactly compute these metrics to Appendix D.

We perform evaluations in three scenarios, similar to \citet{neyshabur2017exploring, keskar2016large}. We compute the 4 metrics on different local minima arising due to (1) a confusion set of varying size in training, (2) different data augmentation schemes, and (3) different batch size.
\begin{itemize}
    \item In Scenario I, we randomly select a subset of 10000 images from CIFAR-10 as the training set and train the DNN with a confusion set consisting of images with random labels. We vary the size of the confusion set so that the resulting local minima generalize differently to the test set while all remain close-to-zero training losses. We consider confusion size of $0$, 1k, 2k, 3k, 4k and 5k. We calculate all metrics based on the sampled 10000 training images.
    \item In Scenario II, we vary the level of data augmentation. We apply horizontal flipping, denoted \code{flip-only}, random cropping from images with 1 pixel padded each side plus flipping, denoted \code{1-crop-f}, random cropping with 4 pixels padded each side plus flipping, denoted \code{4-crop-f} and no data augmentation at all, denoted \code{no-aug}. Under all schemes, the network achieves perfect training accuracy. All the metrics are computed on the un-augmented training set. 
    \item In Scenario III, we vary the batch size. \citet{hoffer2017train} suggests that large batch sizes lead to poor generalization. We consider the batch sizes to be $128$, $256$, $512$ and $1024$.
\end{itemize}
The default values for the 3 variables are confusion size 0, \code{4-crop-f} and batch size 128. For each configuration in each scenario, we train 5 models and report results (average \& standard deviations) of all metrics as well as the test errors (in percentage). For the confusion set experiments, we sample a new training set and a new confusion set every time. In all scenarios, we train the model for 200 epochs with an initial learning rate 0.1, divided by 10 whenever the training loss plateaus. Within each scenario, we find the final training loss very small and very similar across different models and the training accuracy essentially equal to 1, indicating the convergence to local minima.

\begin{figure*}[h!]  
  \centering
  \includegraphics[width=17cm]{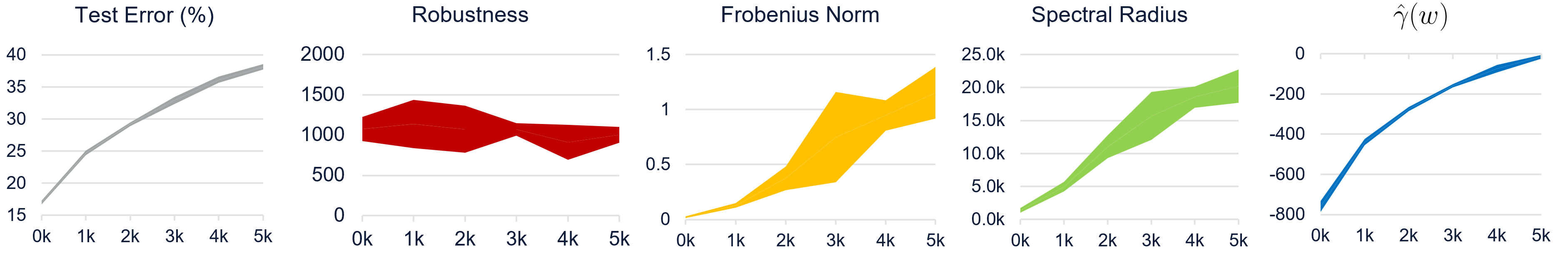}
  \vspace{-5pt}
  \caption{Scenario I: Varied size of the confusion set. 5 models are trained for each size of the confusion set (x-axis). Solid lines are the average result; shaded areas represent the $\pm$ 1 standard deviation (same for Figure \ref{fig:data_aug} and \ref{fig:bs}). A larger confusion set leads to a higher test error, a trend well captured by our metric and the other two; the robustness based metric fails.}
  \label{fig:confusion}
\end{figure*}

\begin{figure*}[h!]  
  \centering
  \includegraphics[width=17cm]{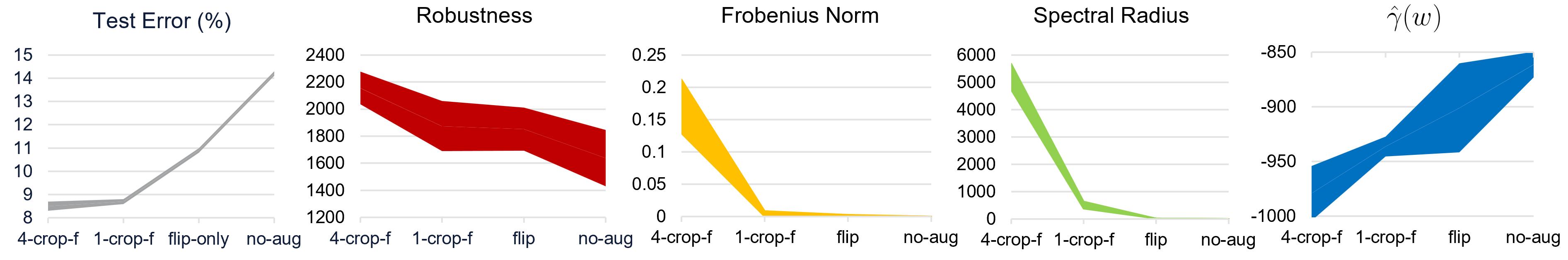}
  \vspace{-5pt}
  \caption{Scenario II: Varied data augmentation schemes. Four different schemes are used. Our metric works well as an indicator of the test error while all the other metrics completely fail.}
  \label{fig:data_aug}
\end{figure*}

\begin{figure*}[h!]  
  \centering
  \includegraphics[width=17cm]{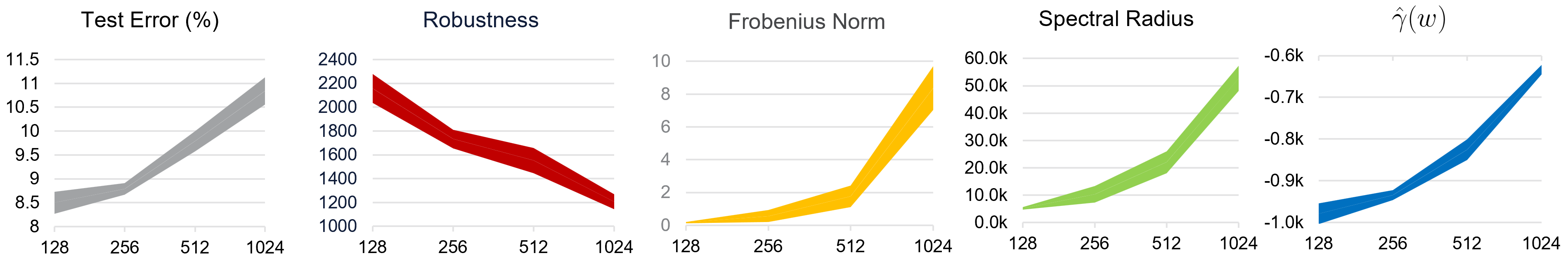}
  \vspace{-5pt}
  \caption{Scenario III: Larger batch size leads to worse generalization, captured by all the metrics except for the robustness based one.}
  \label{fig:bs}
\end{figure*}

The results are in Figure \ref{fig:confusion}, \ref{fig:data_aug} and \ref{fig:bs} for Scenario I, II and III, respectively. Our metric significantly outperforms others and is very effective in capturing the generalization properties, i.e., a lower value of our metric consistently indicates a better generalizable local minimum.  


\begin{table*}[h!]
  \caption{Test error (\%) on CIFAR-10/100. In general, a model with more parameters admits more space for regularization. The representation power of ResNet-20 is too limited for CIFAR-100 (resulting in poor convergence); so we ignore it in our experiments.}
  \vspace{7pt}
  \scriptsize
  \centering
  \setlength{\tabcolsep}{7pt}
  \begin{tabular}{lll|ll|ll|ll}
    \toprule
     & CNN & CNN+reg & WRN-28-2 & WRN-28-2+reg & DenseNet-k12 & DenseNet-k12+reg & ResNet-20 & ResNet-20+reg\\
    \midrule
     CIFAR-10 & 8.52 $\pm$ 0.23 & \textbf{7.55 $\pm$ 0.06} & 5.63 $\pm$ 0.20 & \textbf{5.15 $\pm$ 0.09} & 4.61 $\pm$ 0.08 & \textbf{4.37 $\pm$ 0.06} & 8.50 $\pm$ 0.31 & \textbf{7.89 $\pm$ 0.13}\\
     CIFAR-100 & 31.12 $\pm$ 0.35 & \textbf{29.27 $\pm$ 0.17} & 25.71 $\pm$ 0.24 & \textbf{23.88 $\pm$ 0.13} & 22.54 $\pm$ 0.32 & \textbf{22.23 $\pm$ 0.21} &  - & - \\
    \bottomrule
  \end{tabular}
 \label{tab:reg}
 \end{table*}
 
\begin{table*}[h!]
  \caption{Validation set error (\%) on $128\times 128$ down-sampled ImageNet classification. The better results are bolded.}
  \vspace{7pt}
  \scriptsize
  \centering
  \setlength{\tabcolsep}{7pt}
  \begin{tabular}{llll|llll}
    \toprule
    Top1 Error (\%) & Test & Train & Average Gap & Top5 Error (\%) & Test & Train & Average Gap \\
    \midrule
    WRN-18 & 35.52 $\pm$ 0.11 & 23.67 $\pm$ 2.05 & 11.85 & & 14.27 $\pm$ 0.02 & 7.33 $\pm$ 2.25 & 6.94 \\
    WRN-18+reg & \textbf{34.99 $\pm$ 0.10} & \textbf{24.0 $\pm$ 3.11} & \textbf{10.99} & & \textbf{13.85 $\pm$ 0.05} & \textbf{7.31 $\pm$ 1.07} & \textbf{6.54} \\
    \bottomrule
  \end{tabular}
 \label{tab:reg2}
 \end{table*}
 
\subsection{Experiments on Local Minima Regularization} \label{exp_reg}
We evaluate our regularizer on CIFAR-10, CIFAR-100 and the ImageNet classification task \citep{imagenet_cvpr09}. For CIFAR-10 \& CIFAR-100, we evaluate on four different network architectures including a plain CNN, ResNet-20, Wide ResNet \citep{wideresnet} and DenseNet \citep{huang2017densely}. We use WRN-28-2-B(3,3) from \citet{wideresnet} and the DenseNet-BC-k=12 from \citet{huang2017densely}. 
We evaluate ImageNet classification on WRN-18-1.5 from \citet{wideresnet}.
In specific, we follow \citet{sokolic2017robust} to down-sample all images to $128\times 128$ and apply standard data augmentations.
See Appendix E for architecture and training details.
We denote the four networks as CNN, ResNet-20, WRN-28-2 / WRN-18 and DenseNet-k12, respectively.

For the three hyper-parameters $\alpha, \beta, M$ in our proposed Algorithm \ref{reg_algo}, we find $\alpha$ and $M$ quite robust and manually set $\alpha = 0.0001$, $M = 8$ in all experiments and select $\beta$ by validation via a 45k/5k training data split for each of the network architecture \& dataset pair. In specific, we consider $\beta \in \{1, 5, 10, 20, 30, 40, 50, 75, 100\}$. We keep all the other training hyper-parameters, schemes as well as the setup identical to those in their original paper whenever possible (details in Appendix E). We train 5 separate models for each network-dataset combination on CIFAR-10 and CIFAR-100 and train 3 models for ImageNet. We report the test errors in percentage (mean $\pm$ std.) in Table \ref{tab:reg} and \ref{tab:reg2}, where ``+reg'' indicates training with our regularizer applied. The results demonstrate that our method provides consistent generalization improvement for a wide range of DNNs.

\subsubsection{Time Complexity for Algorithm \ref{reg_algo}}
We benchmark WRN-18 on the down-sampled ImageNet classification dataset with 2 Nvidia 2080 Ti GPUs and a batch size of 128. With parallelization, the average training time per mini-batch is 185.7ms without regularizer applied vs. 285.6ms with regularizer applied.
It only takes around 1.5x longer time per gradient update for Algorithm \ref{reg_algo}.

By ablation study, we find that our regularizer works the best in the mid and late stage of DNN training, e.g., we only use the regularized update after the first learning rate drop in all of our experiments. In the beginning stage where the optimization process is not stable, our regularizer can result in great numerical errors. By only applying Algorithm \ref{reg_algo} during the later stages, the training speed can be further increased by a large margin.

\subsubsection{The Choice of The Optimizer}
As described in Algorithm \ref{reg_algo}, our proposed regularizer is not tied to a specific optimizer. We perform experiments with SGD+Momentum because it is chosen to be used in ResNet, WRN, and DenseNet, helping all of them achieve current or previous state-of-the-art results. Our regularizer aims to find better ``flatter'' minima to improve generalization whereas adaptive optimization methods such as Adam \citep{kingma2014adam} and AdaGrad \citep{duchi2011adaptive} try to boost up convergence, yet usually at the cost of generalizability. Recent works \citep{wilson2017marginal, keskar2017improving} show that adaptive methods generalize worse than SGD+Momentum. In specific, very similar to our setup, \citet{keskar2017improving} demonstrates that SGD+Momentum consistently outperforms the others on ResNet and DenseNet for CIFAR-10 and CIFAR-100. Other approaches that also utilize local curvature to improve SGD, such as the Entropy-SGD \citep{chaudhari2016entropy} mentioned in Sec. \ref{related_work}, have empirical results rather preliminary compared to ours. 

\begin{table}[h!]
  \caption{The proposed metric computed on local minima obtained with or without applying the proposed regularizer. Each entry represents mean $\pm$ std. among 5 runs. Smaller values are bolded.}
  \vspace{7pt}
  \scriptsize
  \centering
  \setlength{\tabcolsep}{4pt}
  \begin{tabular}{l|l|l|l}
    \toprule
     & ResNet-20 & WRN-28-2 & DenseNet-k12 \\
    \midrule
     w/o reg. & -979.3 $\pm$ 22.3 & -689.6 $\pm$ 24.9 & -850.3 $\pm$ 23.5 \\
     with reg. & \textbf{-1138.1 $\pm$ 11.0} & \textbf{-748.7 $\pm$ 21.3} & \textbf{-886.2 $\pm$ 20.5} \\
    \bottomrule
  \end{tabular}
 \label{tab:metric}
 \end{table}

\subsubsection{Generalization Boost As a Result of Better Local Minima}
We perform a sanity check to illustrate that our regularizer indeed induces better local minima characterized by our metric, i.e., our proposed regularizer is consistent with our proposed metric. For ResNet, Wide-ResNet and DenseNet trained on CIFAR-10, we compute the metric on local minima obtained with or without applying the regularizer. 
In specific, our regularizer has an impact on the optimization process, leaving training loss slightly different for models with or without the regularizer.
To ensure our assumption that those local minima have similar close-to-zero training loss, before computing $\hat{\gamma}$ for each model, we normalize and scale the softmax output for each individual training sample. This operation makes comparison between different DNN models robust without changing their underlying behaviors.
Table \ref{tab:metric} shows that the resulting generalization boost aligns with what captured by our metric.

\section{Conclusion and Future Work}
In this paper, we show a bridge between the field of deep learning theory and regularization methods with respect to the generalizability of local minima. We propose a metric that captures the generalization properties of different local minima and provide its theoretical analysis including a generalization bound. We further derive an efficient approximation of the metric and a practical and effective regularizer. Empirical results demonstrate our success in both capturing and improving the generalizability of DNNs. 

Moreover, we find that our proposed regularizer might be further simplified and a dynamic scheduling of the hyper-parameter $\beta$ can provide even more improvement to the generalization performance.
In general, our exploration promises a direction for future work on the regularization and optimization of DNNs.

\paragraph{Acknowledgment}
This work was supported in part by NSF awards CNS-1730158.

\bibliography{example_paper}
\bibliographystyle{icml2020}

\onecolumn
\vskip 0.3in
\appendix
\icmltitle{Appendix}

\section{Proof of Equation 1 in Section 4}
Let us first review the Equation 1 in Section 4:
\begin{equation*}
\mathcal{I}_{\mathcal{S}}(w_0) = \nabla_w^2 \mathcal{L}(\mathcal{S}, w_0) 
= \mathlarger{\mathop{\mathbb{E}}}_{(x, c_x) \sim \mathcal{S}} [\nabla_w \ln p_{w_0}(c_x) \nabla_w \ln p_{w_0}(c_x)^T]
\end{equation*}
To prove this equation, it suffices to prove the following equality: 
\[- \nabla_w^2 \ell\ell_{\mathcal{S}}(w) = \sum_{(x, y) \in \mathcal{S}} \sum_{i=1}^K y_i [\nabla_w \ln p(c_x=i|x;w) \nabla_w \ln p(c_x=i|x;w)^T]\]
For convenience, we change the notation of the local minimum from $w_0$ to $w$ and further denote $p(c_x = i|x; w)$ as $p_w^{x}(i)$. Since $- \nabla_w^2 \ell\ell_{\mathcal{S}}(w) = - \sum_{(x, y) \in \mathcal{S}} \sum_{i=1}^K y_i\ \nabla_w^2 \ln p_w^{x}(i)$, for each $(x, y) \in \mathcal{S}$ and $i \in \{1, 2, ..., K\}$, we have:
\begin{align} \label{eq:a}
    [\nabla_w^2 \ln p_w^{x}(i)]_{j, k} &= \frac{\partial^2}{\partial w_j \partial w_k} \ln p_w^{x}(i) \nonumber\\ 
    &= \frac{\partial}{\partial w_j} \bigg(\frac{\frac{\partial}{\partial w_k} p_w^{x}(i)}{p_w^{x}(i)}\bigg) \nonumber\\
    &= \frac{p_w^{x}(i) \frac{\partial^2}{\partial w_j \partial w_k} p_w^{x}(i)}{p_w^{x}(i)^2} - \frac{\frac{\partial}{\partial w_j}p_w^{x}(i)}{p_w^{x}(i)} \frac{\frac{\partial}{\partial w_k}p_w^{x}(i)}{p_w^{x}(i)} \nonumber\\
    &= \frac{\frac{\partial^2}{\partial w_j \partial w_k} p_w^{x}(i)}{p_w^{x}(i)} - \frac{\partial}{\partial w_j} \ln p_w^{x}(i) \cdot \frac{\partial}{\partial w_k} \ln p_w^{x}(i)
\end{align}
Since $w_0$ is a local minimum of full training accuracy, as described in Section 4, and $y_i = p_w^{x}(i)$ for $i \in \{1, 2, ..., K\}$, when taking the double summation, the first term in Equation \ref{eq:a} becomes:
\[\sum_{(x, y) \in \mathcal{S}} \sum_{i=1}^K \frac{\partial^2}{\partial w_j \partial w_k} p_w^{x}(i) = \frac{\partial^2}{\partial w_j \partial w_k} \sum_{(x, y) \in \mathcal{S}} \sum_{i=1}^K p_w^{x}(i) = \frac{\partial^2}{\partial w_j \partial w_k} N = 0\]
Then it follows that:
\[[\nabla_w^2 \ell\ell_{\mathcal{S}}(w)]_{j,k} = - \sum_{(x, y) \in \mathcal{S}} \sum_{i=1}^K y_i [\nabla_w \ln p_w^{x}(i)\ \nabla_w \ln p_w^{x}(i)^T]_{j,k}\]

\vskip 0.2in
\section{Proof of the Generalization Bound in Section 5.2}
Remind that in Section 5.2 we pick a uniform prior $\mathcal{P}$ over $w \in \mathcal{M}(w_0)$ and pick the posterior $\mathcal{Q}$ of density $q(w) \propto e^{-|\mathcal{L}_0 - \mathcal{L}(\mathcal{S}, w)|}$ with $\mathcal{L}_0 \delequal \mathcal{L}(\mathcal{S}, w_0)$. Then we have the upper bound of the expected generalization loss $\mathop{\mathbb{E}}_{w \sim \mathcal{Q}} [\mathcal{L}(\mathcal{D}, w)]$ in terms of the expected training loss $\mathop{\mathbb{E}}_{w \sim \mathcal{Q}} [\mathcal{L}(\mathcal{S}, w)]$ and $\gamma(w_0)$.

% \begin{thm} \label{main}
% Given $|\mathcal{S}| = N$, $\mathcal{D}$, $\mathcal{L}(\mathcal{S}, w)$ and $\mathcal{L}(\mathcal{D}, w)$ described in Section 3, a local minimum $w_0$, the volume $V$ of $\mathcal{M}(w_0)$ sufficiently small, the Assumption 1 \& 2 satisfied, and $\mathcal{P}, \mathcal{Q}$ defined above, for any $\delta \in (0, 1]$, we have with probability at least $1 - \delta$ that: 
% \begin{equation*}
% \mathop{\mathbb{E}}_{w \sim \mathcal{Q}} [\mathcal{L}(\mathcal{D}, w)] \leq \mathop{\mathbb{E}}_{w \sim \mathcal{Q}} [\mathcal{L}(\mathcal{S}, w)] + 2 \sqrt{ \frac{2 \mathcal{L}_0 + 2 \mathcal{A} + \ln \frac{2N}{\delta}}{N - 1}} \quad
% \textrm{where}\ \mathcal{A} = \frac{W V^{\frac{2}{W}} \pi^{\frac{1}{W}} e^{\red{\gamma(w_0)}/W}}{4\pi e}
% \end{equation*}

% \end{thm}

To prove Theorem 1, let us review the PAC-Bayes Theorem in \citet{mcallester2003simplified}:
\begin{thm} \label{pac_bayes}
For any data distribution $\mathcal{D}$ and a loss function $\mathcal{L}(\cdot, \cdot) \in [0, 1]$, let $\mathcal{L}(\mathcal{D}, w)$ and $\mathcal{L}(\mathcal{S}, w)$ be the expected loss and training loss respectively for the model paramterized by $w$, with the training set $|\mathcal{S}| = N$. For any prior distribution $\mathcal{P}$ with a model class $\mathcal{C}$ as its support, any posterior distribution $\mathcal{Q}$ over $\mathcal{C}$ (not necessarily Bayesian posterior), and for any $\delta \in (0, 1]$, we have with probability at least $1 - \delta$ that:
\[\mathop{\mathbb{E}}_{w \sim \mathcal{Q}} [\mathcal{L}(\mathcal{D}, w)] \leq \mathop{\mathbb{E}}_{w \sim \mathcal{Q}} [\mathcal{L}(\mathcal{S}, w)] + 2 \sqrt{\frac{2 \KL(\mathcal{Q}||\mathcal{P}) + \ln \frac{2N}{\delta}} {N - 1}}\]
\end{thm}

\begin{namedtheorem}[PAC-Bayes (McAllester) ]
For a data distribution $\mathcal{D}$ and a loss $\mathcal{L}(\cdot, \cdot) \in [0, 1]$, let $\mathcal{L}(\mathcal{D}, w)$ and $\mathcal{L}(\mathcal{S}, w)$ be the expected loss and the training loss; the training set $|\mathcal{S}| = N$ is sampled from $\mathcal{D}$. Given arbitrary prior $\mathcal{P}$ and posterior $\mathcal{Q}$ (no need to be Bayesian posterior) supported on a model class $\mathcal{C}$, and for any $\delta > 0$, we have, with probability at least $1 - \delta$, that
\[\mathop{\mathbb{E}}_{w \sim \mathcal{Q}} [\mathcal{L}(\mathcal{D}, w)] \leq \mathop{\mathbb{E}}_{w \sim \mathcal{Q}} [\mathcal{L}(\mathcal{S}, w)] + 2 \sqrt{\frac{2 \KL(\mathcal{Q}||\mathcal{P}) + \ln \frac{2N}{\delta}} {N - 1}}\]
\end{namedtheorem}

As $e^{\gamma(w_0)} = |\mathcal{I}_{\mathcal{S}}(w_0)|$, we can rewrite the generalization bound we want to prove above as:
\begin{equation*} 
\mathop{\mathbb{E}}_{w \sim \mathcal{Q}} [\mathcal{L}(\mathcal{D}, w)] \leq \mathop{\mathbb{E}}_{w \sim \mathcal{Q}} [\mathcal{L}(\mathcal{S}, w)]
+ 2 \sqrt{ \frac{W\cdot V^{2/W} \pi^{1/W} \big\vert \mathcal{I}_{\mathcal{S}}(w_0) \big\vert^{1/W} + 4\pi e \mathcal{L}_0 + 2\pi e \ln \frac{2N}{\delta}}{2\pi e (N - 1)}} 
\end{equation*}

As defined in Section 5.2, given the model class $\mathcal{M}(w_0)$, whose volume is $V$, for the neural network $f_w$, the uniform prior $\mathcal{P}$ attains the probability density function $p(w) = \frac{1}{V}$ for any $w \in ~\mathcal{M}(w_0)$ and the posterior $\mathcal{Q}$ has density $q(w) \propto e^{-|\mathcal{L}(\mathcal{S}, w) - \mathcal{L}_0|}$. Based on Assumption 2 in Section 5.2 and the observed Fisher information $\mathcal{I}_{\mathcal{S}}(w_0)$, especially the Equation 2 derived in Section 4, we have:
\begin{equation*}
\mathcal{L}(\mathcal{S}, w) = \mathcal{L}_0 + \frac{1}{2} (w - w_0)^T \mathcal{I}_{\mathcal{S}}(w_0) (w - w_0)\quad \forall w \in \mathcal{M}(w_0)
\end{equation*}
Denote $\Sigma = [\mathcal{I}_{\mathcal{S}}(w_0)]^{-1} = [\nabla^2_{w}\mathcal{L}(\mathcal{S}, w_0)]^{-1}$. Then $\mathcal{Q}$ is a truncated multivariate Gaussian distribution whose density function $q$ is: 
\begin{align}
q(w; w_0, \Sigma)
 &= \frac{\sqrt{(2\pi)^{-n}|\Sigma|^{-1}} \exp\{-\frac{1}{2} (w - w_0)^T \Sigma^{-1} (w - w_0)\}}{\int_{\mathcal{M}(w_0)} \sqrt{(2\pi)^{-n}|\Sigma|^{-1}} \exp\{-\frac{1}{2} (w - w_0)^T \Sigma^{-1} (w - w_0)\}\ dw} \nonumber \\ 
 &= \frac{\exp\{-\frac{1}{2} (w - w_0)^T \Sigma^{-1} (w - w_0)\}}{\int_{\mathcal{M}(w_0)} \exp\{-\frac{1}{2} (w - w_0)^T \Sigma^{-1} (w - w_0)\}\ dw} \label{eq:3}
\end{align}
Denote the denominator of Equation \ref{eq:3} as $\textbf{Z}$ and define:
\[g(w; w_0, \Sigma) \delequal -\frac{1}{2} (w - w_0)^T \Sigma^{-1} (w - w_0)\} \leq 0\]
Then $q$ can also be written as: 
\[q(w; w_0, \Sigma) = \frac{\exp\{g(w; w_0, \Sigma)\}}{\textbf{Z}}\]
In order to derive a generalization bound in the form of the PAC-Bayes Theorem, it suffices to prove an upper bound of the KL divergence term:
\begin{eqnarray*}
\KL(\mathcal{Q}||\mathcal{P}) & = & \mathop{\mathbb{E}}\limits_{w \sim \mathcal{Q}} \ln \frac{q(w)}{p(w)} \nonumber \\
  & = & -\mathop{\mathbb{E}}\limits_{w \sim \mathcal{Q}} \ln \frac{1}{V} + \mathop{\mathbb{E}}\limits_{w \sim \mathcal{Q}} \ln q(w) \nonumber \\
 & = & \ln V + \mathop{\mathbb{E}}\limits_{w \sim \mathcal{Q}} g(w; w_0, \Sigma) + \ln \frac{1}{\textbf{Z}} \nonumber \\
 & \leq & \ln V + \mathop{\mathbb{E}}\limits_{w \sim \mathcal{Q}} 0 - \ln \bigg(\int_{\mathcal{M}(w_0)} \exp\{g(w; w_0, \Sigma)\}\ dw \bigg) \nonumber \\
 & \leq & \ln V - \ln \bigg(\int_{\mathcal{M}(w_0)} \exp\{-\mathop{\textrm{max}}\limits_{w \in \mathcal{M}(w_0)} \mathcal{L}(\mathcal{S}, w)\}\ dw \bigg) \nonumber \\
 & = & \ln V - \ln \bigg(V \cdot \exp\{-\mathop{\textrm{max}}\limits_{w \in \mathcal{M}(w_0)} \mathcal{L}(\mathcal{S}, w)\}\bigg) \nonumber \\
 & = & \ln V - \ln V + h \quad = \quad h
\end{eqnarray*}
where $h$ is the height of $\mathcal{M}(w_0)$ defined in Section 5.1. For convenience, we shift down $\mathcal{L}(\mathcal{S}, w)$ by $\mathcal{L}_0$ and denote the shifted training loss $\mathcal{L}_0(w) \delequal \mathcal{L}(\mathcal{S}, w) - \mathcal{L}_0$ so that $\mathcal{L}_0(w_0) = 0$. Then
\[\mathcal{L}_0(w) = \frac{1}{2} (w - w_0)^T \Sigma^{-1} (w - w_0)\quad \forall w \in \mathcal{M}(w_0)\]
Furthermore, the following two sets are equivalent
\[\{w \in \mathbb{R}^W: \mathcal{L}(\mathcal{S}, w) = h\} = \{w \in \mathbb{R}^W: \mathcal{L}_0(w) = h - \mathcal{L}_0\}\]
both of which are the $W$-dimensional hyperellipsoid given by the equation $\mathcal{L}_0(w) = h - \mathcal{L}_0$, which can be converted to the standard form for hyperellipsoids as:
\[ (w - w_0)^T \frac{\Sigma^{-1}}{2(h-\mathcal{L}_0)} (w - w_0) = 1\]
The volume enclosed by this hyperellipsoid is exactly the volume of $\mathcal{M}(w_0)$, i.e., $V$; so we have 
\[\frac{\pi^{W/2}}{\Gamma(\frac{W}{2} + 1)} \sqrt{2^W (h-\mathcal{L}_0)^W |\Sigma|} = V\]
Solve for $h$, with the Stirling's approximation for factorial $\displaystyle \Gamma(n + 1) \approx \sqrt{2\pi n} \Big(\frac{n}{e}\Big)^n$, we have
\[h = \mathcal{L}_0 + \frac{\big(V\cdot \Gamma(\frac{W}{2} + 1)\big)^{2/W}}{2\pi \big\vert\Sigma\big\vert^{1/W}} \approx \mathcal{L}_0 + \frac{V^{2/W} \pi^{1/W} W^{(W+1)/W} \big\vert \mathcal{I}_{\mathcal{S}}(w_0) \big\vert^{1/W}}{4\pi e}\]
where $\Gamma(\cdot)$ denotes the Gamma function. Notice that for modern DNNs we have $W \gg 1$, and so $\displaystyle W^{\frac{W+1}{W}} \approx W$. We finally can derive the generalization bound in the form of the PAC-Bayes Theorem as:
\[\mathop{\mathbb{E}}_{w \sim \mathcal{Q}} [\mathcal{L}(\mathcal{D}, w)] \leq \mathop{\mathbb{E}}_{w \sim \mathcal{Q}} [\mathcal{L}(\mathcal{S}, w)] + 2 \sqrt{ \frac{W\cdot V^{2/W} \pi^{1/W} \big\vert \mathcal{I}_{\mathcal{S}}(w_0) \big\vert^{1/W} + 4\pi e \mathcal{L}_0 + 2\pi e \ln \frac{2N}{\delta}}{2\pi e (N - 1)}}\]
% \paragraph{Remark} The bound given here based on the expected training loss is slightly tighter than the one given in the main paper, since we can upper bound $\mathop{\mathbb{E}}_{w \sim \mathcal{Q}} [\mathcal{L}(\mathcal{S}, w)]$ by:
% \[\mathop{\mathbb{E}}_{w \sim \mathcal{Q}} [\mathcal{L}(\mathcal{S}, w)] \leq \mathop{\mathbb{E}}_{w \sim \mathcal{Q}} [\mathop{\textrm{max}}\limits_{w \in \mathcal{M}(w_0)} \mathcal{L}(\mathcal{S}, w)] = h\]

% \mathop{\mathbb{E}}_{w' \sim \mathcal{Q}} [\mathcal{L}(\mathcal{D}, w')] \leq \mathop{\mathbb{E}}_{w'
%  \sim \mathcal{Q}} [\mathcal{L}(\mathcal{S}, w')] + 2 \sqrt{ \frac{WV^{2/W} \pi^{1/W} \exp\{\color{red}\gamma(w)\color{black}/W\} + 4\pi e \mathcal{L}_w + 2\pi e \ln \frac{2N}{\delta}}{2\pi e (N - 1)}}

\vskip 0.2in
\section{Derivation of Equation 6 in Section 5.3} \label{app_appro}
First, let us present the well-known theorem in linear algebra that relates the eigenvalues of a matrix to those of its sub-matrices. 

\begin{thm} \label{eig_thm}
Given an $n \times n$ real symmetric matrix A with eigenvalues $\lambda_1 \leq ... \leq \lambda_n$, for any $k < n$ denote its principal sub-matrix as $B$ obtained from removing $n - k$ rows and columns from $A$. Let $\nu_1 \leq ... \leq \nu_k$ be the eigenvalues of $B$. Then for any $1 \leq r \leq k$, we have $\lambda_r \leq \nu_r \leq \lambda_{r+n-k}$.
\end{thm}

Let $\{\nu_n\}_{n=1}^{N'}$ be the eigenvalues of $\frac{1}{W} \xi^t(w_0)$, which is a $N' \times N'$ sub-matrix of $\mathcal{I}_{\mathcal{S'}}(w_0)$; then
\[\widehat{\gamma}(w_0) = \frac{1}{T} \sum_{t = 1}^{T} \ln \big\vert\xi^t(w_0)\big\vert = \frac{1}{T} \sum_{t = 1}^{T} \ln\big\vert W \cdot \frac{1}{W} \xi^t(w_0)\big\vert = N' \ln W + \frac{1}{T} \sum_{t = 1}^{T} \sum_{n = 1}^{N'} \ln\nu_n\]
Theorem \ref{eig_thm} gives the relation between $\nu_n$ and $\lambda_n$, defined above and in Section 5.3 as the $n^{\rm{th}}$ smallest eigenvalues of $\frac{1}{W} \xi^t(w_0)$ and that of $\mathcal{I}_{\mathcal{S'}}(w_0)$, respectively. For sufficiently large $N'$, we can use $\nu_n$ to approximate $\lambda_n$, which ignores the eigenvalues of  $\mathcal{I}_{\mathcal{S'}}(w_0)$ larger than $\lambda_{N'}$. This is reasonable when estimating $\gamma(w_0)$, since in general the majority of the eigenvalues of the Hessian for DNNs are close to zero with only a few large ``outliers'', and so the smallest eigenvalues are the dominant terms in $\gamma(w_0)$ \citep{pennington2018spectrum, sagun2017empirical, karakida2019universal}. A specific bound of the eigenvalues remains an open question, though. In short, we have $\sum_{n = 1}^{N'} \nu_n \approx \sum_{n = 1}^{N'} \lambda'_n$ and consequently:
\begin{align*}
    \frac{W}{N'}\widehat{\gamma}(w_0) + W \ln \frac{1}{W} &= \frac{W}{N'}\widehat{\gamma}(w_0) - W \ln W \\
    &= \frac{W}{N'} \Big(\widehat{\gamma}(w_0) - N' \ln W\Big) \\
    &= \frac{1}{T} \sum_{t = 1}^{T} \frac{W}{N'} \sum_{n = 1}^{N'} \ln \nu_n \\
    &\approx \frac{1}{T} \sum_{t = 1}^{T} \frac{W}{N'} \sum_{n = 1}^{N'} \ln\lambda_n'
\end{align*}
Finally we we have
\[\lim_{T \rightarrow \infty}\frac{1}{T} \sum_{t = 1}^{T} \frac{W}{N'} \sum_{n = 1}^{N'} \ln\lambda_n' = \gamma(w_0)\]

\vskip 0.2in
\section{Details of Calculating the Metrics in Section 7.1} \label{app_metrics}
For the following three metrics, we apply estimation by sampling a subset $\mathcal{S}^t$ from the full training set $\mathcal{S}$ for $T$ times and averaging the results.
\begin{itemize}
  \item Frobenius norm: $\big\Vert \nabla^2_w \mathcal{L}(\mathcal{S}, w)\big\Vert_F^2$ 
  \item Spectral radius: $\rho( \nabla^2_w \mathcal{L}(\mathcal{S}, w))$
  \item Ours: $\widehat{\gamma}(w) =\frac{1}{T} \sum_{t = 1}^{T} \ln |\xi(\mathcal{S}^t, w_0)|$
\end{itemize}

For the Frobenius norm based metric, from Equation 1 \& 2 in Section 4 we have:
\[\big\Vert \nabla^2_w \mathcal{L}(\mathcal{S}, w)\big\Vert_F^2 = \big\Vert \mathcal{I}_{\mathcal{S}}(w)\big\Vert_F^2 = \frac{1}{N}\sum_{(x, y) \in \mathcal{S}} \sum_{i=1}^K \Big\Vert\big(\nabla_{w}[\boldsymbol{\ell}_x(w_0)]_i\big) \big(\nabla_{w}[\boldsymbol{\ell}_x(w_0)]_i\big)^T\Big\Vert^2_F\]
We define $\mathbf{y} = \argmax(y)$. 
Similar to Equation 4 in Section 5.3, we approximate $y$ by $\tilde{y}$ and so
\[\big\Vert \nabla^2_w \mathcal{L}(\mathcal{S}, w)\big\Vert_F^2 \approx \frac{1}{N}\sum_{(x, y) \in \mathcal{S}} \Big\Vert\big(\nabla_{w}[\boldsymbol{\ell}_x(w_0)]_{\mathbf{y}}\big) \big(\nabla_{w}[\boldsymbol{\ell}_x(w_0)]_{\mathbf{y}}\big)^T\Big\Vert^2_F\]
Summing over the entire Hessian matrix is too expensive as there are $W \times W \times N$ entries in total. We therefore estimate the quantity by first sampling a subset $\mathcal{S}^t \subset \mathcal{S}$ and then sampling 100,000 entries of $\big(\nabla_{w}[\boldsymbol{\ell}_x(w_0)]_{\mathbf{y}}\big) \big(\nabla_{w}[\boldsymbol{\ell}_x(w_0)]_{\mathbf{y}}\big)^T$. We perform the estimation $T$ times and average the results, similar to the approach when computing $\widehat{\gamma}(w)$.

Also by Equation 2 and the approximation in Equation 4, the spectral radius of Hessian is equivalent to the squared spectral norm of $1/\sqrt{N}\mathbf{J}_{w}[\tilde{\mathcal{L}}(\mathcal{S}, w)]$. We also perform estimation (with irrelevant scaling constants dropped) by sampling $\mathcal{S}^t$ for $T$ times, i.e., via $\frac{1}{T}\sum_{t} \big\Vert\mathbf{J}_{w}[\tilde{\mathcal{L}}(\mathcal{S}^t, w)]\big\Vert_2^2$.

Furthermore, in all our experiments that involves samplings $\mathcal{S}^t$, we set $|\mathcal{S}^t| = N' = T = 100$.

\vskip 0.2in
\section{Architecture And Training Details in Section 7}
Architecture details are as below
\begin{itemize}
    \item The plain CNN is a 6-layer convolutional neural network similar to the baseline in \citet{lee2016generalizing} yet without the ``mlpconv'' layers (resulting in a much fewer number of parameters). Specifically, the 6 layers has numbers of filters as $\{64, 64, 128, 128, 192, 192\}$. We use $3 \times 3$ kernel size and ReLU as the activation function. After the second and the fourth convolutional layer we insert a $2 \times 2$ max pooling operation. After the last convolutional layer, we apply a global average pooling before the final softmax classifier. 
    \item For ResNet-20, WRN-28-2-B(3,3), WRN-18-1.5 and DenseNet-BC-k=12, we use the same architecture as in their original papers, respectively.
\end{itemize}

The training details are
\begin{itemize}
    \item For the plain CNN, we initialize the weights according to the scheme in \citet{he2016deep} and apply l2 regularization of a coefficient $0.0001$. We perform standard data augmentation, the one denoted \code{4-crop-f} in Section 7.1. We use stochastic gradient descent with Nesterov momentum set to 0.9 and a batch size of 128. We train 200 epochs in total with the learning rate initially set to 0.01 and then divided by 10 at epoch 100 and 150.
    \item For ResNet-20, WRN-28-2-B(3,3), WRN-18-1.5 and DenseNet-BC-k=12, we use the same hyper-parameters, training schemes, data augmentation schemes, optimization methods, etc., as those in their original papers, respectively. An exception is that for WRN-18-1.5 on ImageNet, we first resize all training images to $128\times 128$, and then apply random crop (of size $114 \times 114$), horizontal flip and standard color jittering together with mean channels subtraction as in \citet{he2016deep}. We adopt single crop (central crop) testing for the down-sampled $128\times 128$ validation images.
\end{itemize}

\end{document}